\documentclass{article}
\usepackage[utf8]{inputenc}
\usepackage[T1]{fontenc}
\usepackage{geometry}
\usepackage{graphicx}
\usepackage{amsmath, amssymb}
\usepackage{natbib}
\usepackage{booktabs}
\usepackage{tabularx}
\usepackage{hyperref}
\usepackage{microtype}  
\usepackage{float}
\usepackage{lineno}  

\title{Can We Test Consciousness Theories on AI? Ablations, Markers, and Robustness}
\author{Yin Jun Phua\\
\textit{Institute of Science Tokyo}\\
\texttt{phua@comp.isct.ac.jp}}
\date{\today}

\begin{document}


\maketitle

\begin{abstract}
\noindent The search for reliable indicators of consciousness has fragmented into competing theoretical camps (Global Workspace Theory (GWT), Integrated Information Theory (IIT), and Higher-Order Theories (HOT)), each proposing distinct neural signatures. We adopt a synthetic neuro-phenomenology approach: constructing artificial agents that embody these mechanisms to test their functional consequences through precise architectural ablations impossible in biological systems. Across three experiments, we report dissociations suggesting these theories describe complementary functional layers rather than competing accounts. In Experiment~1, a no-rewire Self-Model lesion abolishes metacognitive calibration while preserving first-order task performance, yielding a synthetic blindsight analogue consistent with HOT predictions. In Experiment~2, workspace capacity proves causally necessary for information access: a complete workspace lesion produces qualitative collapse in access-related markers, while partial reductions show graded degradation, consistent with GWT's ignition framework. In Experiment~3, we uncover a broadcast-amplification effect: GWT-style broadcasting amplifies internal noise, creating extreme fragility. The B2 agent family is robust to the same latent perturbation; this robustness persists in a Self-Model-off / workspace-read control, cautioning against attributing the effect solely to $z_{\text{self}}$ compression. We also report an explicit negative result: raw perturbational complexity (PCI-A) \emph{decreases} under the workspace bottleneck, cautioning against naive transfer of IIT-adjacent proxies to engineered agents. A simple composite of markers (GBI~+~$\Delta$PCI~+~AUROC) predicts OOD fragility. These results suggest a hierarchical design principle: GWT provides broadcast \emph{capacity}, while HOT provides \emph{quality control}. We emphasize that our agents are not conscious; they are reference implementations for testing functional predictions of consciousness theories.
\end{abstract}

\section{Introduction}

The science of consciousness faces a validation crisis. While theories like Global Workspace Theory (GWT) \citep{baars1988cognitive, dehaene2011experimental}, Integrated Information Theory (IIT) \citep{tononi2004information, oizumi2014phenomenology}, and Higher-Order Theories (HOT) \citep{rosenthal2005consciousness, lau2011empirical} propose distinct neural signatures of awareness (ranging from non-linear ``ignition'' \citep{dehaene2014consciousness} and causal integration ($\Phi$) \citep{oizumi2014phenomenology} to metacognitive self-monitoring \citep{fleming2010relating}), testing them in biological systems is constrained by the inability to perform precise architectural ablations. We propose \textit{synthetic neuro-phenomenology}: building agents that explicitly embody these theoretical commitments to test their functional utility.

This approach addresses a gap in both neuroscience and artificial intelligence. For neuroscience, synthetic agents offer a "perfect model organism" where every neuron, weight, and activation is observable and manipulable, allowing for causal tests that are ethically or technically impossible in biological brains. For AI, consciousness theories offer a potential roadmap for overcoming the brittleness of current systems \citep{bengio2017consciousness}. While deep learning has achieved broad success, modern models often lack the robustness, adaptability, and self-monitoring capabilities characteristic of biological intelligence. By implementing architectural priors derived from consciousness science (such as global broadcasting \citep{goyal2021recurrent} and self-modeling), we aim to determine whether these mechanisms confer the same functional advantages to artificial agents that they are hypothesized to provide to biological organisms.

However, prior attempts to model consciousness in silico have often fallen short of systematic validation. Many implementations rely on loose metaphorical translations of neural theories, where a simple feedback loop is labeled ``recurrent processing'' or a central variable is called a ``global workspace'' without establishing the functional necessity of these features. Furthermore, existing studies typically focus on single theories in isolation (building a ``GWT agent'' \citep{goyal2021recurrent} or a ``HOT-inspired Self-Model agent'' \citep{wilterson2021attention}), ignoring the potential interactions and dependencies between these mechanisms. Few studies subject these architectures to ``adversarial phenomenology'': actively trying to break the system via lesions, noise, and deceptive signaling to distinguish genuine architectural advantages from mere task-specific overfitting.

In this work, we address these shortfalls by subjecting our agents to a battery of stress tests. We move beyond merely confirming theory predictions to stress-testing them in a controlled synthetic environment. To do this, we designed agents with reinforcement-learning-style architectures—trained here via supervised behavior cloning from an oracle (used only to generate training targets) to isolate architectural effects from optimization confounds—that incorporate specific architectural features proposed by consciousness theories. Specifically, we built agents with (1) a \textbf{global workspace} to broadcast information between modules (representing GWT), (2) a \textbf{Self-Model} to monitor internal states (representing HOT), and (3) \textbf{complexity probes} to measure information integration (inspired by IIT). Because these are software agents, we can perform manipulations that are impossible in biology: we apply \textbf{causal lesions} (selectively breaking parts of the ``brain''), \textbf{noise titration} (injecting precise amounts of interference to test robustness), and \textbf{adversarial reward structures} (creating environments that incentivize the agent to ``lie'' to itself).

Our findings reveal a functional hierarchy rather than competing theories:
\begin{enumerate}
    \item \textbf{Action Without Metacognition (Synthetic Blindsight):} Consistent with HOT predictions, we find that in this agent family, ablating the Self-Model selectively abolishes metacognitive calibration (``knowing that you know'') while leaving first-order task performance largely intact, yielding a synthetic analogue of blindsight where agents can \textit{do} without \textit{knowing}.
    \item \textbf{Bus-Off Discontinuity in Access Markers:} We observe an ignition-analogous discontinuity in this finite, discrete-capacity workspace: across the capacity settings we test, access markers remain similar across the tested non-zero capacities but collapse when the workspace is fully lesioned (bus-off)\footnote{Here, we use ``ignition-analogous'' in a loose access sense: an abrupt change between tested \emph{discrete} capacity settings in a finite system. We do \emph{not} claim a thermodynamic-limit phase transition, power-law scaling, or a precisely estimated critical point.}. Reducing workspace capacity does not gradually dim access within the tested levels; instead, a complete workspace lesion produces a qualitative collapse in access-related markers.
    \item \textbf{Broadcast-Amplification vs.\ Noise-Attenuation:} Contrary to the intuition that broadcasting creates robustness \citep{baars1988cognitive, bengio2017consciousness}, we find that a global workspace alone exhibits a \emph{broadcast-amplification} effect: it broadcasts information but also amplifies noise, making the system fragile even under standard regularization and noise augmentation. In our implementation, the B2 agent family remains robust in the same perturbation regime (L75 $> 0.50$), including in a Self-Model-off / workspace-read control; we therefore treat noise-attenuation as an empirical robustness property of the B2 family rather than as a bottleneck-only effect.
    \item \textbf{A Transfer Constraint on IIT-Adjacent Proxies:} Raw perturbational complexity (PCI-A) is \emph{inverted} under the workspace bottleneck: the GWT agent shows lower trajectory complexity than baselines, contrary to a naive ``more integrated $\Rightarrow$ higher PCI'' expectation. This explicit negative result constrains how PCI-style measures transfer to engineered agents and suggests that absolute algorithmic complexity is not an architecture-invariant proxy for integration.
\end{enumerate}

These results suggest that consciousness markers are not redundant. GWT provides the \textit{capacity} for information exchange, while HOT mechanisms provide the \textit{control} necessary to stabilize that exchange. Neither alone is sufficient for robust agency.

\textbf{Scope and Limitations.} Throughout this work, we focus strictly on \textit{Access Consciousness} (information availability for report and action) rather than phenomenal consciousness or subjective experience. Our agents are not conscious; they are reference implementations of proposed mechanisms. Our claims concern functional correlates within these synthetic systems: whether certain architectural features are necessary or sufficient for particular behavioral and representational signatures. We make no claims about whether these features would produce phenomenal experience in biological or artificial systems. Because we evaluate up to $n=20$ seeds per condition (Experiments~1--3; seed as the unit of analysis), our statistical claims are best viewed as proof-of-concept evidence; we highlight large qualitative effects and report uncertainty estimates for transparency. Some targeted sweeps and attribution controls are pilots ($n=5$ seeds) and are labeled as such.

\section{Background}
\label{sec:background}

\subsection{Neuroscience \& Consciousness Concepts}
\label{sec:neuro_concepts}
To bridge the gap between biological theory and synthetic implementation, we clarify key neuroscientific terms used throughout this work.

\paragraph{Type-1 vs. Type-2 Performance.}
In psychophysics, \textbf{Type-1 performance} refers to the objective accuracy of a subject's decision (e.g., "is the stimulus left or right?"). \textbf{Type-2 performance} refers to the subject's ability to monitor their own accuracy, typically measured via confidence ratings (e.g., "how sure are you?"). A key signature of conscious awareness is the correlation between Type-1 and Type-2 performance (metacognitive sensitivity). Conditions like \textit{blindsight} \citep{weiskrantz1986blindsight} demonstrate that high Type-1 performance can exist without Type-2 awareness, a dissociation we model in Experiment~1.

\paragraph{Ignition and Broadcasting.}
Global Workspace Theory posits that conscious access involves a non-linear ``ignition'' event, an abrupt thresholding dynamic in which a stimulus overcomes a barrier to trigger sustained, widespread reverberation across fronto-parietal networks \citep{dehaene2006conscious}. This "broadcasting" makes the information available to other cognitive processes. We model the capacity-limited broadcast aspect via a bottlenecked bus architecture; our implementation tests capacity constraints on information maintenance and broadcast, rather than competitive access gating.

\paragraph{Backward Masking and Visual Working Memory.}
\textit{Backward masking} is an experimental technique in which a brief target stimulus is followed by a ``mask'' (a second stimulus) that interferes with conscious perception of the target. The \textit{stimulus-onset asynchrony} (SOA), the time between target and mask onset, determines whether the target reaches awareness. This paradigm probes the temporal dynamics of conscious access. \textit{Visual working memory} refers to the capacity-limited system that maintains visual information over short delays (typically 3--4 items in humans \citep{luck1997capacity, cowan2001magical}). We use masking-like interference and capacity manipulations to test analogous dynamics in our synthetic agents.

\paragraph{Wagering Paradigms.}
A \textit{wagering} or \textit{post-decision wagering} paradigm asks subjects to bet on the correctness of their own decisions \citep{persaud2007post}. High wagers on correct trials and low wagers on incorrect trials indicate good metacognitive calibration. An \textit{opt-out} option (declining to wager) provides an additional measure of subjective uncertainty. We adapt this paradigm to assess whether our agents can calibrate confidence to their own accuracy.

\paragraph{TMS--EEG and PCI.}
The Perturbational Complexity Index (PCI) \citep{casali2013theoretically} is measured by perturbing the brain with transcranial magnetic stimulation (TMS, a non-invasive technique that induces neural activity via magnetic pulses) and recording the resulting spatiotemporal dynamics with electroencephalography (EEG, scalp recordings of electrical brain activity). PCI quantifies the algorithmic complexity of the brain's response: conscious states produce complex, differentiated responses, while unconscious states (e.g., deep sleep, anesthesia) produce simpler, more stereotyped responses. Our PCI-A proxy applies an analogous perturbation-and-compress logic to synthetic agent trajectories.

\subsection{Related Work}

Our work builds on and extends several research threads that have previously remained largely separate.

\paragraph{Consciousness-Inspired AI Architectures.}
Several recent efforts have sought to implement consciousness-related mechanisms in artificial systems. \citet{goyal2021recurrent} introduced recurrent independent mechanisms with attention-based communication, inspired by GWT's broadcast architecture. \citet{vanrullen2021deep} explored attention-gated global workspaces in vision models. \citet{wilterson2021attention} developed HOT-inspired architectures with explicit self-monitoring. However, these works typically focus on single theories in isolation and demonstrate performance gains without systematic ablation or stress-testing to establish causal necessity. Our contribution is to (1)~implement multiple theoretical mechanisms within a unified architecture, (2)~systematically lesion each component to establish necessity, and (3)~subject the system to adversarial perturbations that distinguish genuine robustness from task-specific~fitting.

\paragraph{Metacognition and Calibration in Machine Learning.}
The machine learning literature has extensively studied uncertainty quantification and calibrated confidence \citep{guo2017calibration}. Selective prediction frameworks \citep{geifman2017selective} allow models to abstain when uncertain, and recent work connects these ideas to human-like metacognition \citep{raileanu2020automatic}. Our Type-2 AUROC metric derives from this literature but is deployed here to test HOT-specific predictions: that metacognitive calibration requires a higher-order self-representation rather than emerging automatically from first-order competence. The synthetic blindsight dissociation we report (preserved task performance with abolished metacognition) provides a controlled test case for these ideas.

\paragraph{Robustness and Noise in Neural Networks.}
The fragility of deep networks to adversarial perturbations \citep{szegedy2014intriguing} and distribution shift \citep{hendrycks2019benchmarking} is well documented. Prior work has explored noise injection during training for regularization \citep{srivastava2014dropout} and studied the geometry of learned representations under perturbation \citep{mahloujifar2019curse}. Our broadcast-amplification finding connects to this literature but focuses on \emph{latent} perturbations (noise in the workspace, not inputs) and demonstrates that broadcasting alone amplifies rather than filters noise. The B2 family's robustness in the same latent-noise regime suggests that additional architectural structure and training objectives can provide complementary robustness mechanisms beyond input-level defenses.

\paragraph{Metacognition in Reinforcement Learning.}
While uncertainty quantification in deep learning is well-established \citep{guo2017calibration}, recent work has begun to explore the functional role of metacognition in reinforcement learning agents. \citet{anderson2020metacognitive} demonstrated that metacognitive monitoring can enhance perturbation tolerance in RL agents, a finding consistent with our noise-attenuation hypothesis. Similarly, \citet{salemgarcia2023linking} linked confidence biases to fundamental RL processes, suggesting that miscalibration is not merely a post-hoc artifact but can actively shape learning dynamics. Our work extends these perspectives by causally dissociating first-order performance from second-order confidence via architectural lesions, providing a synthetic testbed for theories that posit hierarchical relationships between task competence and metacognitive awareness \citep{fleming2023metacognition}.

\section{From Theories to Synthetic Indicators}

Having established the relevant neuroscientific terminology and experimental paradigms, we now turn to the core challenge of this work: translation. To move beyond loose analogies between brains and machines, we must operationalize theoretical constructs into precise, measurable indicators within our synthetic agents. We selected three dominant clusters of consciousness theories (HOT, GWT, and IIT) because they address distinct functional facets of consciousness that are often conflated: (1) \textbf{Monitoring} (reflexive awareness), (2) \textbf{Access} (information availability), and (3) \textbf{Integration} (phenomenological unity). By testing these pillars individually, we can determine whether they are redundant or complementary.

\paragraph{Higher-Order Theories (HOT).}
HOT suggests that a state is conscious only when the system represents itself as being in that state \citep{rosenthal2005consciousness, lau2011empirical}. This corresponds to the \textbf{Monitoring} dimension: access alone is insufficient for confidence, error correction, and a ``personal'' perspective. We implement this via a dedicated \textbf{Self-Model} ($z_{\text{self}}$) that compresses the workspace state into a summary latent. The indicator is \textbf{Metacognitive Calibration} (Type-2 AUROC), measuring the system's ability to ``know that it knows'' \citep{fleming2010relating}. We test this in Experiment~1 to dissociate \emph{doing} (first-order performance) from \emph{knowing} (second-order awareness).

\paragraph{Global Workspace Theory (GWT).}
GWT posits that consciousness arises from a global broadcast mechanism where information from specialized modules is selected and amplified (``ignited'') for system-wide access \citep{baars1988cognitive, dehaene2011experimental}. This corresponds to the \textbf{Access} dimension: without global broadcasting, there is no ``content'' to be conscious of. In our agents, we operationalize this as a bottlenecked bus architecture. The key indicators are \textbf{Ignition}, a sudden, non-linear jump in workspace activation, and the \textbf{Global Broadcast Index (GBI)}, quantifying the extent of information sharing across the system. We test this in Experiment~2 to establish the causal substrate of information access.

\paragraph{Integrated Information Theory (IIT).}
IIT argues that consciousness corresponds to the capacity of a system to integrate information, measured as $\Phi$ \citep{tononi2004information, oizumi2014phenomenology}. This corresponds to the \textbf{Integration} dimension: consciousness is characteristically unified and differentiated.

We do not compute $\Phi$ directly. Instead, we include a PCI-inspired perturbational complexity probe \citep{casali2013theoretically} as a \emph{transfer test}: if ``PCI-like'' trajectory complexity were a broadly architecture-invariant proxy for IIT-style integration, then architectures that enforce more global coupling and causal interdependence should tend to exhibit higher perturbation-evoked complexity.

However, in engineered agents this expectation is not guaranteed \citep{shwartz2017opening, ansuini2019intrinsic}. In particular, a global workspace can behave as a \emph{bottleneck} and \emph{organizer}: by forcing many downstream computations to depend on a shared, low-rank broadcast state, it can increase redundancy and push dynamics onto a lower-dimensional manifold. Under such conditions, stronger architectural ``integration'' (in the everyday engineering sense of tighter coupling) can plausibly produce \emph{lower} Lempel--Ziv compressibility-based trajectory complexity.

Accordingly, we treat raw PCI-A as a falsifiable probe rather than a confirmatory marker, and we interpret deviations (including inversions) as informative constraints on how PCI-style measures do or do not transfer to artificial systems. We report both raw PCI-A (absolute post-perturbation trajectory complexity) and a contrastive measure, \textbf{$\Delta$PCI} (correct minus incorrect trial complexity), which aims to partially control for the fact that in these agents, ``complexity'' can be inflated by unstructured failure/noise. We include this in Experiment~3 to test which aspects of PCI-style proxies behave monotonically with architectural changes in this synthetic setting.

These indicators form a distinct profile for each theory, summarized in Table~\ref{tab:indicators}.

\begin{table}[h]
\centering
\caption{Mapping theoretical clusters to synthetic indicators and experiments.}
\label{tab:indicators}
\begin{tabularx}{\textwidth}{cllX}
\toprule
\textbf{Exp.} & \textbf{Theory Cluster} & \textbf{Intended Property} & \textbf{Synthetic Indicator} \\
\midrule
E1 & HOT / Self-Model & Metacognitive access & Type-2 AUROC, $z_{\text{self}}$ lesion \\
\hline
E2 & GWT & Global access, ignition & GBI, Ignition Sharpness \\
\hline
E3 & Complexity probe (IIT-adjacent) & Richness, differentiation & PCI-A (raw), $\Delta$PCI \\
\hline
E3 & Composite & Joint ``conscious-like'' profile & Reduced composite (GBI + $\Delta$PCI); full CTS adds AUROC \\
\bottomrule
\end{tabularx}
\end{table}

\section{Methods}

\subsection{Architectures}

We compared a hierarchy of agent architectures, all sharing a common backbone but differing in their high-level cognitive modules.

\paragraph{Common Backbone.}
All agents use a convolutional encoder (3-layer CNN) to process the $\mathbf{O}_{\text{grid}} \in \mathbb{N}^{7 \times 7 \times 3}$ input, producing a 151-dimensional observation embedding.

\paragraph{Feedforward Baseline (A0).}
A purely reactive agent consisting of an MLP (Multi-Layer Perceptron) policy head attached directly to the visual encoder. It lacks memory or internal state, serving as a control for tasks requiring temporal integration.

\paragraph{Recurrent Baseline (A1).}
A standard recurrent agent where the observation embedding feeds into a Gated Recurrent Unit (GRU) with 64 hidden units. The policy head reads directly from the GRU hidden state. This represents the "status quo" of deep reinforcement learning agents—capable of memory but lacking explicit workspace or Self-Model structures.

\paragraph{GWT Agent (B1): The Global Workspace.}
This agent implements a capacity-limited broadcast mechanism.
\begin{itemize}
    \item \textbf{Workspace Bus:} A set of $K$ memory slots (default $K=4$, dimension $D=16$).
    \item \textbf{Write Mechanism:} Recurrent state features are projected to a write vector and stored in the fixed-capacity buffer. At each step, the write vector is placed into the first available empty slot; if all slots are occupied, a fixed slot (slot 0) is overwritten.\footnote{We use fixed-slot overwrite rather than true FIFO (evicting the oldest entry) for implementation simplicity. In our short-episode tasks (32 steps) with sparse cue-write events, the workspace rarely fills completely, so the fallback policy seldom activates. When it does, the flattened-readout architecture ensures all slot contents remain accessible to the policy regardless of which slot was overwritten. This design choice therefore does not affect the validity of our capacity-limitation experiments.} This deterministic allocation tests \emph{capacity limitation} and \emph{broadcast}, not competitive attention-based access gating.
    \item \textbf{Broadcast (Read):} The content of all slots is aggregated and broadcast back to the policy head. We use a flattened-readout MLP that concatenates all workspace slots into a single vector and projects through a multi-layer network to produce action logits.
    \item \textbf{Capacity Control:} We can manipulate the number of active slots $K$ (full capacity $K=4$, intermediate reductions, and $K=0$ (bus-off) for a complete lesion).
\end{itemize}

\paragraph{Self-Model Agent (B2; HOT-inspired).}
This agent extends B1 by adding a higher-order monitoring system.
\begin{itemize}
    \item \textbf{Self-Model ($z_{\text{self}}$):} A secondary module that does not directly see the world. Instead, it observes the \textit{workspace broadcast} (a $D$-dim summary), the \textit{recurrent carrier state}, and the \textit{agent's own intended actions} (policy logits).
    \item \textbf{Compression:} It compresses this high-dimensional input (89 dims: 64-d recurrent carrier + 16-d workspace broadcast + 7 action logits + 2 uncertainty scalars) into a tiny latent bottleneck ($z_{\text{self}}$, dim=64, effectively lower due to regularization).
    \item \textbf{Policy Routing:} The confidence head (for wagering) reads \textit{exclusively} from $z_{\text{self}}$. The action policy routing varies by experiment: in Experiment~1 (metacognition), the action policy reads directly from the workspace (not $z_{\text{self}}$) so that Self-Model lesions preserve first-order performance while abolishing metacognition. In Experiment~3 (noise robustness), we evaluate both configurations: a ``workspace-read'' control where the action policy bypasses $z_{\text{self}}$, and the standard B2 where the action policy reads from $z_{\text{self}}$.
\end{itemize}

\paragraph{Baseline Variants for Disentangling Contributions.}
Two controls help separate the roles of broadcast and self-modeling. (i) We evaluate B2 with the Self-Model enabled under a complete workspace lesion (bus-off; $K=0$), which tests whether a monitor can produce spurious metacognition when its monitored substrate is absent; Type-2 AUROC collapses to chance (see Experiment~1). (ii) We include compression and pathway controls: blind ablations (noise/permutation) and a PCA baseline matched to the effective dimensionality of $z_{\text{self}}$ to show that metacognitive information depends on the learned content of $z_{\text{self}}$, not merely on the presence of a bottleneck.

\textbf{Denoising attribution baselines (Experiment~3 controls).} To tighten attribution for the robustness effects in Experiment~3, we additionally evaluate two baseline families. First, a ``HOT-only'' agent (A1 + $z_{\text{self}}$ without any workspace bus) tests whether a Self-Model bottleneck can improve robustness in the absence of a broadcast substrate. Because this agent has no workspace bus, it cannot be evaluated under the strong-lesion cue routing (cues-to-workspace-only), so cues remain visible in its observation stream; we probe robustness by injecting noise into the recurrent carrier state (post-GRU hidden state) rather than into workspace slots. Second, we include matched ``BC-only'' compressor controls that replace $z_{\text{self}}$ with a 64-d bottleneck applied directly to the flattened workspace state (linear projection, small MLP, and fixed random projection). These controls distinguish a Self-Model-specific robustness gain from a generic benefit of adding a low-dimensional bottleneck; results are summarized in Figure~\ref{fig:e3_denoise_controls}.
 
 \subsection{Environments}

All experiments were conducted within customized \textbf{MiniGrid} environments \citep{chevalier2018minigrid}, a standard 2D gridworld framework where agents perceive the world through a local view ($7 \times 7$ grid) and take discrete actions (move forward, turn left, turn right, pick up, drop, toggle, done). The observation space consists of a symbolic tensor $\mathbf{O}_{\text{grid}} \in \mathbb{N}^{7 \times 7 \times 3}$ (object ID, color, state) plus a task-specific vector.

\paragraph{Wagering Paradigm (Experiment~1).}
To assess metacognition, we implemented a wagering task within the MiniGrid framework. The agent navigates to a ``stimulus zone'' where a transient cue (encoded via the standard MiniGrid observation channels) appears with variable salience (simulating perceptual uncertainty). The agent must then navigate to one of two ``response zones'' to indicate its classification.
\begin{itemize}
    \item \textbf{Task:} Classify which of two cue types ($s \in \{0, 1\}$) was presented, where cue salience is degraded by observation noise.
    \item \textbf{Wager:} On 50\% of trials, a ``wager flag'' appears after the decision, prompting the agent to output a continuous confidence signal $c \in [0, 1]$ via a dedicated action channel.
    \item \textbf{Reward Structure:} High confidence ($c > 0.5$) counts as a ``bet'' (+1 reward for correct, -1 for incorrect). Low confidence ($c \le 0.5$) counts as an ``opt-out'' (0 reward). This incentivizes the agent to align its internal confidence $c$ with its objective probability of correctness $P(\text{correct})$.
\end{itemize}

\paragraph{Dual-Task Paradigm (Experiment~2).}
To test GWT predictions, we used a \texttt{MiniGrid-Empty-5x5-v0} variant with a dual-task structure. Each episode (max 32 steps) involves:
\begin{enumerate}
    \item \textbf{Encoding Phase:} Agents observe two transient cues: a ``secondary'' cue (color) and a ``primary'' cue (object type).
    \item \textbf{Delay Phase:} Agents must navigate the empty room while maintaining both cues in working memory.
    \item \textbf{Report Phase 1 (Early):} A specific ``report flag'' signal triggers the agent to output the secondary cue.
    \item \textbf{Report Phase 2 (Late):} A second flag triggers the agent to output the primary cue.
\end{enumerate}
We employed a ``strong-lesion'' wiring: cue information was routed \textit{exclusively} through the workspace bus, while the standard recurrent encoder was blind to these specific cue channels. This ensures the workspace is the sole causal pathway for task-relevant information.

\paragraph{Latent Noise Stress Test (Experiment~3).}
To evaluate robustness on the dual-task paradigm (Experiment~2), we introduced a Latent Workspace Perturbation protocol. Instead of adding noise to the pixel observations (which traditional CNNs can easily filter), we injected Gaussian noise $\mathcal{N}(0, \sigma^2)$ directly into the agent's internal workspace slots immediately \textit{after} the write operation but \textit{before} the read operation. For recurrence-only controls without a workspace (HOT-only), we analogously inject Gaussian noise into the recurrent carrier state (post-GRU hidden state) immediately after the recurrent update and before downstream readouts.
This ``brain stimulation'' approach tests the system's stability against internal degradation. We measured the \textbf{L75}: the noise level $\sigma$ at which report-step decision accuracy drops to 75\%.\footnote{We use ``L75'' rather than ``LD50'' to avoid confusion with the toxicology convention where LD50 denotes the dose causing 50\% lethality. Our threshold uses 75\% accuracy (not 50\% accuracy).} This effectively quantifies the ``tolerance margin'' of the cognitive architecture.

\subsection{Training}

Agents were trained using \textbf{Behavior Cloning (BC)} from an expert oracle to ensure baseline competence. We use BC as an \emph{optimization control}: it reduces RL-specific confounds (exploration failures, reward sparsity, and training instability) so that differences under lesions and perturbations can be attributed to architectural manipulations rather than optimization dynamics. The oracle is used \emph{only} to generate action targets during training; it is never consulted at inference. All evaluations are autonomous rollouts of the learned policy, and our central results are established via \emph{post-training} causal interventions (lesions, capacity ablations, and latent workspace noise) applied to trained agents. This framing makes our claims conditional: given agents that reach baseline first-order competence, which components are causally necessary for access- and metacognition-related signatures?

\subsubsection{Training Procedure}

\paragraph{Dataset Generation.}
We generated training data using a hard-coded algorithmic solver (the ``Oracle'') that has perfect knowledge of the environment dynamics and task rules. For each experiment, we collected oracle demonstrations per agent type and seed (see Table~\ref{tab:hyperparameters} for episode counts by experiment). Each episode consists of up to 32 environment steps. The Oracle's action at each step was recorded as the target distribution for behavior cloning.

\paragraph{Loss Function.}
We minimized the Kullback-Leibler (KL) divergence between the agent's predicted action distribution $\pi_\theta(a|s)$ and the oracle's action distribution $\pi^*(a|s)$:
\[ \mathcal{L}_{\text{BC}} = \mathbb{E}_{s \sim \mathcal{D}} \left[ D_{\text{KL}}(\pi^*(a|s) \| \pi_\theta(a|s)) \right] \]
where $\mathcal{D}$ is the dataset of oracle demonstrations.

\paragraph{Optimization.}
Training proceeded for the number of episodes specified in Table~\ref{tab:hyperparameters} (with one optimizer update per step) using the Adam optimizer \citep{kingma2014adam} with default momentum parameters ($\beta_1 = 0.9$, $\beta_2 = 0.999$). The learning rate was set to $10^{-3}$ for all agents. We used online training (processing each episode immediately rather than batching across episodes) to maintain temporal coherence of recurrent states. No learning rate scheduling or warm-up was employed.

\paragraph{Convergence Criteria.}
Training was run for a fixed number of episodes rather than early stopping, as behavior cloning from a deterministic oracle converges reliably. We verified convergence by confirming that agents achieved $>$95\% oracle-action agreement on held-out episodes (per-step argmax match) by the end of training. Seeds that failed to reach this imitation threshold were flagged and excluded from analysis.

\subsubsection{Stabilization Techniques}

The dual-task paradigm creates two challenges for training: (1) temporal class imbalance, where report steps constitute only $\sim$3\% of all steps, and (2) the need for the Self-Model to learn meaningful representations without direct supervision. We addressed these with targeted interventions:

\begin{itemize}
    \item \textbf{Report-Step Reweighting:} Report steps are rare but critical for evaluation. To prevent these sparse signals from being dominated by navigation steps, we upweighted the KL loss on report steps by a factor of 3.0. Additionally, we applied class-specific weighting: the ``primary cue'' action (correct on only $\sim$5\% of steps) received 20$\times$ higher weight than the ``secondary cue'' action in the loss computation.
    
    \item \textbf{Metacognitive Supervision (B2 agents; Experiments~1 and~3):} For B2 agents, we added a supervised auxiliary loss training the confidence head to predict binary correctness on report steps: $\mathcal{L}_{\text{meta}} = \text{BCE}(\hat{c}, \mathbf{1}_{\text{correct}})$, where $\hat{c}$ is the agent's confidence output. Here, $\mathbf{1}_{\text{correct}}$ is defined by the agent's own \emph{intended} report decision (argmax of its policy logits) being correct, as assessed against the oracle-provided ground-truth label at that report step. This trains calibration on the agent's \emph{own} failure modes (``will my imminent decision be correct?'') rather than on predicting oracle behavior. This places the Self-Model ($z_{\text{self}}$) directly in the gradient path for metacognitive judgments.
    
    \item \textbf{Stimulus Auxiliary Loss (Experiment~1 only):} To ensure the workspace maintains stimulus information throughout the delay period, we added an auxiliary loss training a linear probe on the workspace state to predict stimulus presence at every step, not just report steps. This provides dense gradients that prevent the workspace memory trace from fading before the critical report moment.
\end{itemize}

\subsubsection{Validation Strategy}

We employed a simple held-out validation approach:

\begin{itemize}
    \item \textbf{Train/Test Split:} After training on the specified number of oracle episodes (4,000 for E1/E2; 8,000 for E3; see Table~\ref{tab:hyperparameters}), we evaluated agents on held-out independently sampled episodes with different random seeds (Experiment~2 capacity sweeps use 32 episodes per seed; robustness titrations use 50 episodes per noise level). These held-out episodes used the same task structure but different initial conditions and cue assignments.
    
    \item \textbf{Metrics:} Validation performance was assessed via oracle-action agreement (imitation accuracy) and task-level scores such as conjunction accuracy (both report phases correct) and report-window decision accuracy. We required $>$95\% oracle-action agreement on held-out data to consider training successful.
    
    \item \textbf{No Hyperparameter Tuning on Test Data:} All hyperparameters were fixed before evaluation. The evaluation episodes used for reporting results (Experiments 1--3) were never used during hyperparameter selection.
\end{itemize}

\subsubsection{Hyperparameter Selection}

Hyperparameters were selected through a combination of theory-driven priors and limited manual search:

\begin{itemize}
    \item \textbf{Architecture:} Workspace capacity ($K=4$ slots) and dimension ($D=16$) were chosen to match estimates of human visual working memory capacity \citep{luck1997capacity, cowan2001magical}. The Self-Model latent dimension (64) was set to allow sufficient representational capacity while remaining no larger than the workspace state ($K \times D = 64$).
    
    \item \textbf{Learning Rate:} We tested $\{10^{-4}, 10^{-3}, 10^{-2}\}$ on a single seed and selected $10^{-3}$ based on fastest convergence without instability. This choice was fixed for all subsequent experiments.
    
    \item \textbf{Loss Weights:} The report-step reweighting factor (3.0) and class weight (20.0) were selected to approximately balance the contribution of rare and common step types to the total loss. We did not extensively tune these values.
    
    \item \textbf{No Automated Search:} We did not employ grid search, random search, or Bayesian optimization. The limited hyperparameter exploration reflects our goal of testing architectural hypotheses rather than maximizing benchmark performance.
\end{itemize}

Table~\ref{tab:hyperparameters} summarizes the key hyperparameters used across experiments.

\begin{table}[h]
\centering
\caption{Training hyperparameters by experiment. All experiments used Adam optimizer with default momentum.}
\label{tab:hyperparameters}
\begin{tabular}{lcccc}
\toprule
\textbf{Parameter} & \textbf{E1 (HOT)} & \textbf{E2 (GWT)} & \textbf{E3 (Triangulation)} \\
\midrule
Training episodes & 4,000 & 4,000 & 8,000 \\
Steps per episode & 32 & 32 & 32 \\
Learning rate & $10^{-3}$ & $10^{-3}$ & $10^{-3}$ \\
Workspace slots $K$ & 4 & 4 (varied) & 4 \\
Workspace dim $D$ & 16 & 16 & 16 \\
Self-Model dim & 64 & --- & 64 \\
Report step weight & 3.0 & 3.0 & 3.0 \\
Primary class weight & 20.0 & 20.0 & 20.0 \\
Meta-loss coefficient & 1.0 & --- & 1.0 \\
Random seeds & 20 & 20 & 20 \\
\bottomrule
\end{tabular}
\end{table}

\subsection{Indicator Computation}

\paragraph{Global Broadcast Index (GBI).}
We quantify how broadly information is shared across workspace slots using a participation-style coefficient. At each time step $t$, we treat each workspace slot as a node and define an adjacency matrix from slot similarity:
\[ A_t(i,j) = \mathbf{w}_{t,i}^\top \mathbf{w}_{t,j}, \quad A_t(i,i)=0 \]
where $\mathbf{w}_{t,i}$ is the activation vector of slot $i$. For each slot $i$, we compute a participation coefficient
\[ P_i(t) = 1 - \sum_{j \neq i} \left(\frac{A_t(i,j)}{\sum_{j' \neq i} A_t(i,j') + \varepsilon}\right)^2. \]
GBI is the mean of $P_i(t)$ across slots and time within the decision window. Higher GBI indicates that each slot couples to multiple other slots (broad broadcast), while low GBI indicates narrow coupling/fragmentation.

\paragraph{Ignition Sharpness.}
We detected ``ignition'' events by looking for sudden jumps in the total magnitude of workspace activation. We computed the temporal derivative of the activation norm and measured the peak steepness. A sharp peak indicates an abrupt ignition-like transition; a flat trace indicates gradual or failed access.

\paragraph{Type-2 AUROC (Metacognition).}
We used the Area Under the Receiver Operating Characteristic curve for Type-2 judgments.
\begin{itemize}
    \item \textbf{Input:} For every trial, we record the agent's binary correctness (0 or 1) and its continuous confidence wager $c \in [0, 1]$.
    \item \textbf{Analysis:} We treat confidence as a predictor of correctness. We sweep a threshold $t$ from 0 to 1. For each $t$, we calculate the True Positive Rate (proportion of correct trials with $c > t$) and False Positive Rate (proportion of incorrect trials with $c > t$).
    \item \textbf{Result:} The area under this curve represents the probability that the agent assigns higher confidence to a correct trial than an incorrect one. 0.5 is chance (no insight); 1.0 is perfect insight.
\end{itemize}

\paragraph{Perturbational Complexity (PCI-A, $\Delta$PCI).}
To probe an IIT-adjacent integration proxy, we used a perturbation approach inspired by the Perturbational Complexity Index (PCI).
\begin{enumerate}
    \item \textbf{Perturb:} We inject a pulse of noise into the workspace.
    \item \textbf{Measure:} We record the subsequent trajectory of workspace states.
    \item \textbf{Compress:} We binarize the trajectory and compress it using the Lempel--Ziv algorithm (gzip-style compression).
    \item \textbf{Calculate:} The file size of the compressed trajectory indicates its algorithmic complexity; we report this absolute score as raw PCI-A. Because absolute PCI-A can be inflated by unstructured failure/noise in engineered agents, we also report a within-agent contrast, $\Delta$PCI $= \text{PCI-A}_{\text{correct}} - \text{PCI-A}_{\text{incorrect}}$, which asks whether perturbational complexity is \emph{selectively} higher on successful trials.
\end{enumerate}

\paragraph{No-Report Signature ($\Delta$NRS).}
To characterize whether stimulus information is present in the agent's internal state during the delay period, even when explicit report is blocked, we fit a linear decoder on internal feature vectors from blocked-report steps to predict the binary stimulus label. We score the decoder with AUC and subtract a shuffled-label baseline computed with the same cross-validation splits:
\[ \Delta\text{NRS} = \text{AUC}_{\text{true}} - \mathbb{E}[\text{AUC}_{\text{shuffle}}]. \]
A positive $\Delta$NRS indicates above-baseline decodability; values near zero indicate little or no decodable signal, and negative values indicate below-baseline decoding due to finite-sample noise. In the bus-off condition, the workspace broadcast is identically zero, so $\Delta$NRS reflects what can be decoded from the remaining trunk state alone.

\subsection{Ablations and Manipulations}

We employed precise causal interventions:
\begin{itemize}
    \item \textbf{No-rewire Self-Model lesion (Experiment~1):} We evaluate the same trained checkpoint twice: intact and with the Self-Model removed (so $z_{\text{self}}$ is replaced by zeros), while leaving the policy pathway unchanged. We use checkpoints whose policy reads the workspace directly (not exclusively $z_{\text{self}}$) so that first-order task performance remains intact under the lesion, allowing a clean Type-1/Type-2 dissociation test.
    \item \textbf{Capacity Ablation (Experiment~2):} We varied workspace capacity. For primary behavioral analyses we report three capacities (4, 2, or 0 slots; Figure~\ref{fig:e2_results}). For the no-report signature ($\Delta$NRS), we also ran a 5-level sweep of \texttt{workspace\_capacity\_scale} $\in \{0, 0.25, 0.5, 0.75, 1.0\}$ (0--4 effective slots) to test for a bus-on/bus-off discontinuity.
\end{itemize}

\section{Results}

\subsection{Experiment 1 (HOT): Metacognitive Calibration Depends on the Self-Model}

\subsubsection{Motivation}
Higher-Order Theories (HOT) of consciousness posit that a system becomes conscious of a state only when it represents itself as being in that state \citep{rosenthal2005consciousness, lau2011empirical}. In artificial agents, this implies that effective metacognitive monitoring (``knowing that you know'') requires a dedicated architectural component that re-represents the system's internal state. We selected this angle to test the \textbf{Monitoring} dimension of consciousness, specifically to dissociate first-order \emph{doing} (task competence) from second-order \emph{knowing} (metacognitive awareness), a distinction central to phenomena like blindsight.

In Experiment~1, we operationalize these predictions by equipping a global workspace agent with an explicit \textit{Self-Model} ($z_{\text{self}}$) that compresses internal states into a latent summary. We test three key hypotheses:
\begin{enumerate}
    \item \textbf{Metacognitive Necessity:} Accurate second-order confidence judgments (metacognition) require the Self-Model's re-representation of the workspace state.
    \item \textbf{Causal Specificity:} Ablating the Self-Model should degrade metacognitive calibration (Type-2 AUROC) without impairing first-order task performance, demonstrating a double dissociation.
    \item \textbf{Informational Content:} The Self-Model must actively compress and filter task-relevant information (e.g., uncertainty), rather than acting as a passive passthrough or random projection.
\end{enumerate}

\subsubsection{Results}

\begin{figure}[htbp]
\centering
\begin{tabular}{cc}
\includegraphics[width=0.45\textwidth]{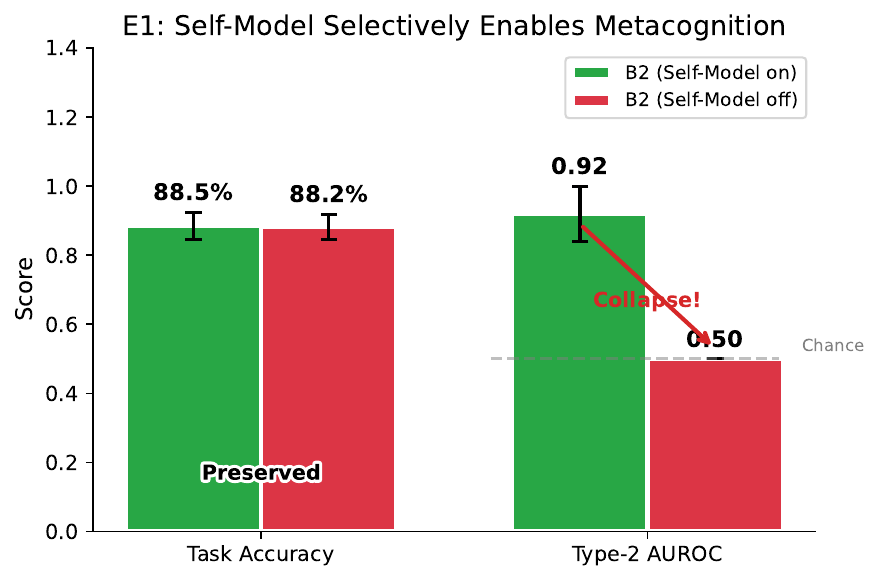} &
\includegraphics[width=0.45\textwidth]{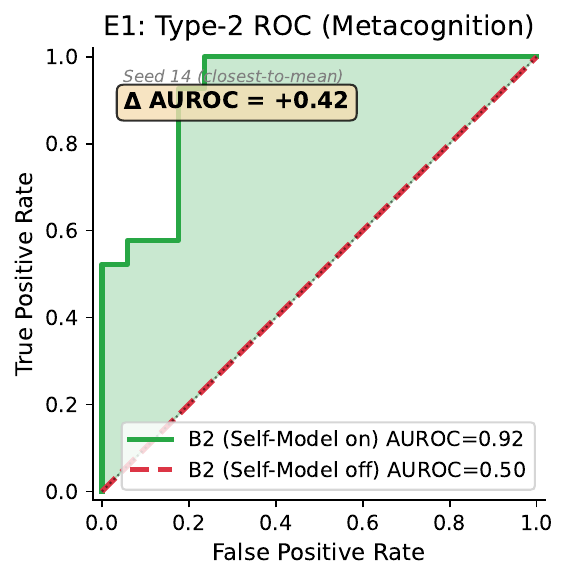} \\
(a) & (b) \\
\includegraphics[width=0.45\textwidth]{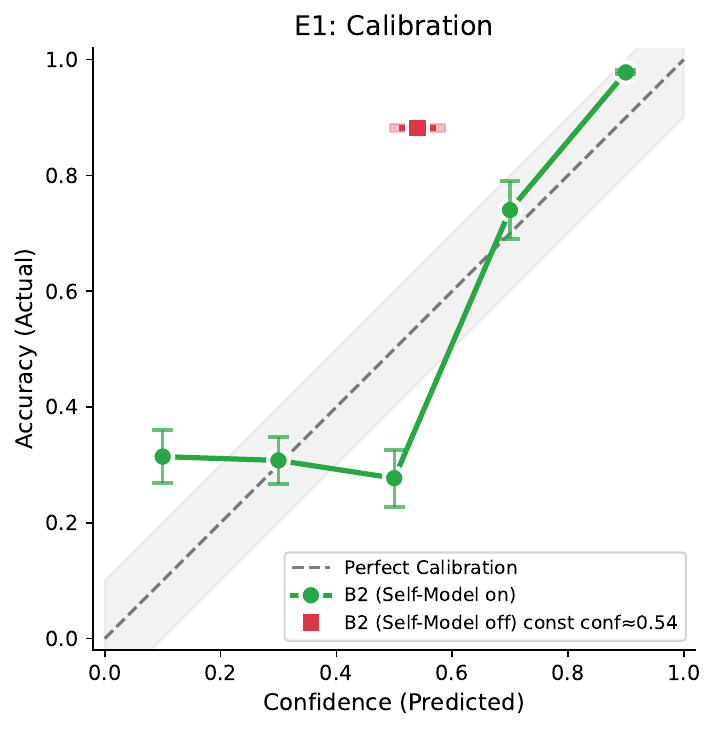} &
\includegraphics[width=0.45\textwidth]{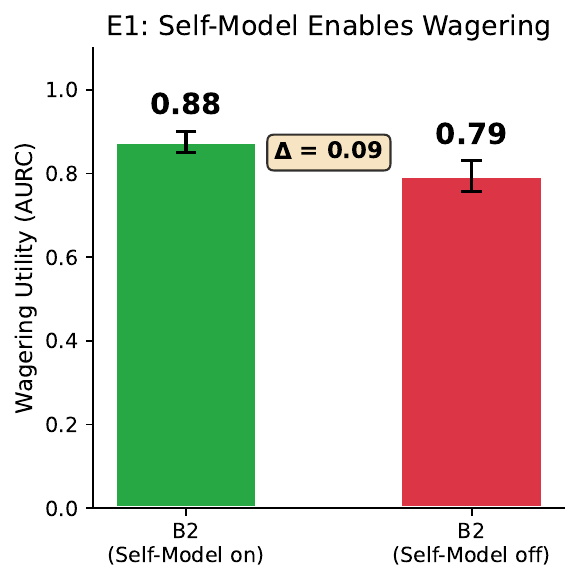} \\
(c) & (d)
\end{tabular}
\caption{\textbf{Metacognitive calibration depends on the Self-Model.} 
(a) First-order task accuracy (left) is preserved under Self-Model ablation, while Type-2 AUROC (right) collapses to chance. Error bars show $\pm 1$ SD across 20 seeds.
(b) Type-2 ROC curves: self-on (green) achieves AUROC = 0.92; self-removed (red) falls on the diagonal (AUROC = 0.50). Shaded region shows metacognitive advantage.
(c) Calibration plot: self-on shows a positive confidence--accuracy relationship; self-removed outputs constant confidence ($\approx 0.54$).
(d) Wagering utility (AURC): self-on achieves higher selective accuracy through calibrated confidence.
}
\label{fig:e1_results}
\end{figure}

Figure~\ref{fig:e1_results} presents the main findings. The Self-Model ablation produced a dissociation between first-order task performance and second-order metacognitive sensitivity.

\paragraph{First-Order Performance is Preserved.}
Task accuracy was largely unchanged between conditions. Across $n = 20$ random seeds, the self-on model achieved $88.5\% \pm 4.0\%$ accuracy (mean $\pm$ SD), while the self-removed model achieved $88.2\% \pm 3.7\%$ accuracy, a difference of only $0.3$ percentage points (Figure~\ref{fig:e1_results}a, left bars). This confirms that the no-rewire Self-Model lesion preserves first-order task performance while ablating metacognitive signal content.

\paragraph{Type-2 AUROC Collapses to Chance.}
Conversely, metacognitive sensitivity was reduced by Self-Model removal. The self-on condition achieved a mean Type-2 AUROC of $0.92$ ($n = 20$ seeds; no-rewire evaluation), indicating well-above-chance metacognitive calibration. The self-removed condition sat exactly at chance (AUROC $= 0.50$ for all seeds), yielding a Self-Model effect of $\Delta\text{AUROC} = +0.42$ (95\% CI $[0.38, 0.45]$). The ROC curves (Figure~\ref{fig:e1_results}b) visualize this collapse: the self-on curve lies well above the diagonal, while the self-removed condition falls on the diagonal (AUROC $= 0.50$).

To put these numbers in context: an AUROC of $0.92$ means the agent's confidence is higher on correct trials 92\% of the time (vs.\ 50\% by chance). The collapse to $0.50$ under ablation indicates confidence ratings became uninformative about performance. This pattern mirrors human blindsight, where patients perform visual discriminations above chance while lacking subjective awareness, and thus any basis for confidence judgments.

\paragraph{Confidence Becomes Constant Without Self-Model.}
Examination of confidence distributions elucidated the mechanism underlying the AUROC collapse: in the self-removed condition, the confidence head outputs a near-constant value ($0.54 \pm 0.02$). Without input from $z_{\text{self}}$, the confidence head has no information to modulate its output (Figure~\ref{fig:e1_results}c).

\paragraph{Workspace Lesion Produces Chance-Level Confidence.}
A distinct question is how the intact Self-Model behaves when the global workspace itself is lesioned ($K=0$; bus-off). One might worry that removing the monitored substrate could yield spuriously high confidence (``nothing contradicts me''), rather than low/flat confidence (``nothing to monitor''). We tested this directly by evaluating B2 with the Self-Model enabled under a bus-off lesion (\texttt{workspace\_capacity\_scale} $= 0.0$). Metacognitive sensitivity collapsed to near chance: Type-2 AUROC $= 0.492 \pm 0.058$ across $n=20$ seeds (mean $\pm$ SD), with a higher opt-out rate (skip fraction $= 0.136 \pm 0.128$). By contrast, at full workspace capacity ($K=4$; bus scale $1.0$), metacognitive calibration is high (Type-2 AUROC $= 0.92$ in the main no-rewire evaluation; Figure~\ref{fig:e1_results}b). Under the seed-level aggregation used for this workspace-lesion control (imputing AUROC $=0.50$ when AUROC is undefined due to perfect first-order performance in a few seeds), this baseline is Type-2 AUROC $= 0.856 \pm 0.180$ with skip fraction $= 0.044 \pm 0.052$. Thus, lesioning the workspace does not produce ``uninhibited'' high confidence; it produces an effectively uninformative confidence signal consistent with the absence of a monitorable global state.

\paragraph{Control Analyses: Blind Ablation and Compression.}
Blind ablations (noise/permutation) also resulted in AUROC collapsing to $\approx 0.50$, confirming that the specific \textit{information content} of $z_{\text{self}}$ is necessary, not merely the presence of an active pathway. Furthermore, analysis of $z_{\text{self}}$ revealed an effective dimensionality of $\approx 2.0$ (compressed from 89 input dimensions: 64+16+7+2), representing a $\sim$98\% reduction in dimensionality. This compression is functionally significant: the Self-Model distills the high-dimensional workspace state into just two independent axes of variation, yet these two dimensions carry sufficient information to support metacognitive judgments. Despite this $\sim$98\% compression, $z_{\text{self}}$ outperformed a PCA baseline in predicting correctness ($+5.1\%$ AUROC advantage), indicating that the Self-Model performs task-relevant filtering rather than merely preserving the directions of maximum variance. In other words, the Self-Model learns \emph{what matters} for metacognition, not just what varies most.

\paragraph{Deep Dive: Workspace Collapse and Passthrough.}
We conducted a deep dive into failure modes. In one outlier seed where the self-on agent failed to achieve calibration (AUROC $\approx 0.5$), analysis revealed that the underlying workspace slots contained no information about correctness (Cohen's $d \approx 0$). A linear probe on the workspace bus outperformed the Self-Model in this case, suggesting the Self-Model acts as a non-linear filter: when the workspace signal is strong, $z_{\text{self}}$ amplifies it for metacognition; when the signal is weak, $z_{\text{self}}$ collapses, effectively destroying the information. This reinforces the hierarchical dependency: the Self-Model cannot read what the workspace has not written.

We next ask whether the underlying workspace capacity itself determines information access, independent of higher-order monitoring.

\subsection{Experiment 2 (GWT): Workspace Capacity Determines Information Access}

\subsubsection{Motivation}
Global Workspace Theory (GWT) posits that conscious access corresponds to a non-linear ``ignition'' event in which information becomes globally available to distributed modular systems via a capacity-limited central resource \citep{baars1988cognitive, dehaene2011experimental}. While neural correlates of ignition are well documented in biological brains \citep{mashour2020conscious}, causal evidence linking workspace capacity, ignition dynamics, and behavioral information access remains sparse. We selected this angle to test the \textbf{Access} dimension of consciousness: determining whether the workspace is merely a passive correlate or the \emph{causal substrate} of information integration and broadcast.

In Experiment~2, we operationalize GWT's core predictions in an artificial agent to test whether:
\begin{enumerate}
    \item \textbf{Causal Necessity:} We test whether access to information for multi-step tasks depends causally on workspace capacity.
    \item \textbf{Ignition Dynamics:} Successful access is marked by a sudden, non-linear increase in workspace activation (``ignition'') and global sharing (broadcast).
    \item \textbf{Ablation Sensitivity:} Reducing workspace capacity degrades both the neural signature (ignition/broadcast) and the behavioral outcome (task accuracy) in tandem.
\end{enumerate}

\subsubsection{Results}

\begin{figure}[htbp]
\centering
\begin{tabular}{cc}
\includegraphics[width=0.45\textwidth]{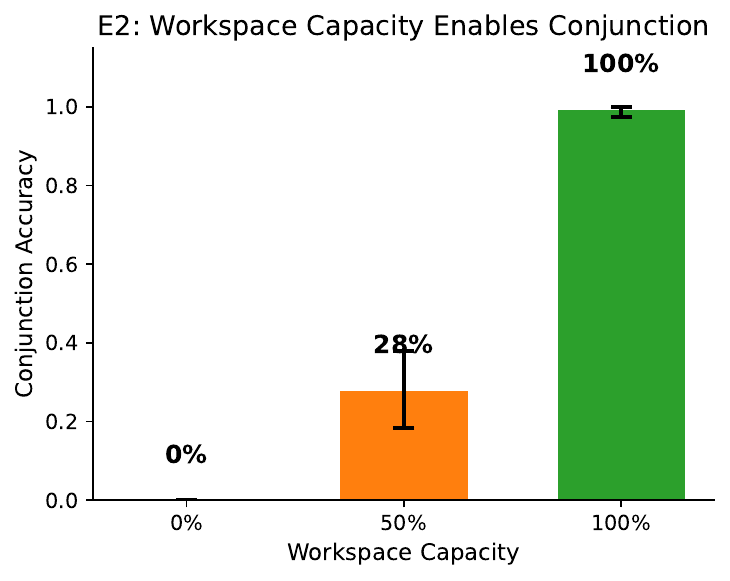} &
\includegraphics[width=0.45\textwidth]{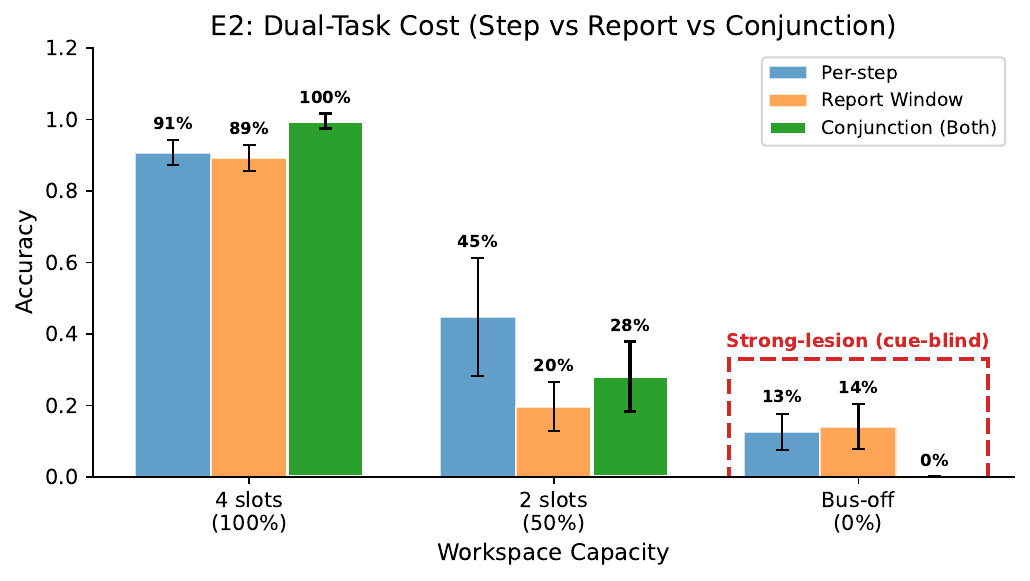} \\
(a) & (b) \\
\multicolumn{2}{c}{\includegraphics[width=0.7\textwidth]{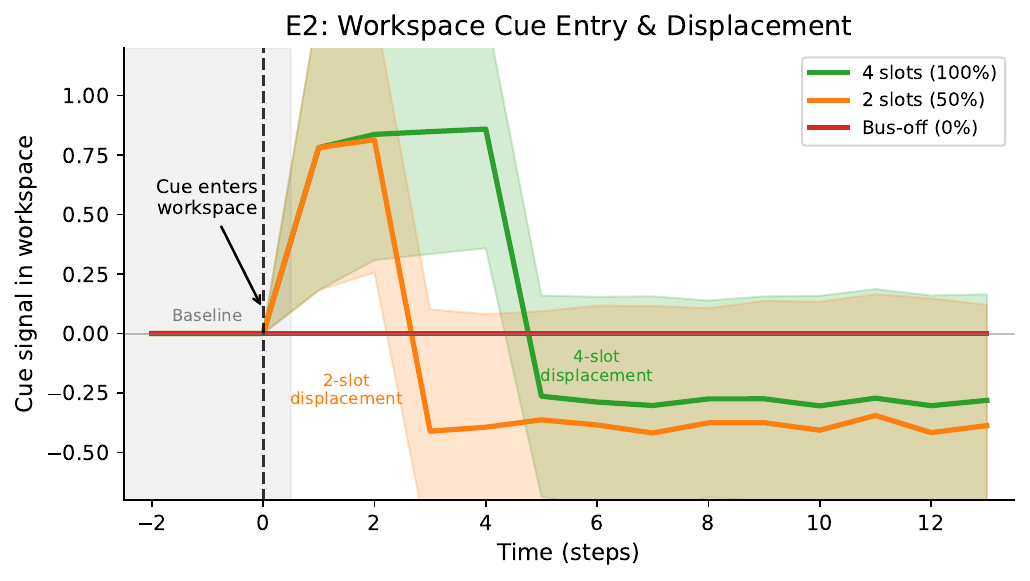}} \\
\multicolumn{2}{c}{(c)}
\end{tabular}
\caption{\textbf{Workspace capacity determines information access.}
(a) Conjunction accuracy by workspace capacity ($n = 20$ seeds; mean $\pm$ SD): $99.5\% \pm 2.0\%$ at 4 slots, $28.1\% \pm 9.7\%$ at 2 slots, $0.0\% \pm 0.0\%$ at bus-off. Full vs. half capacity: Welch's $t$-test $p < 10^{-16}$ (seed-level), Hedges' $g = 9.70$. Baseline agents (A0, A1) achieve near-chance under strong-lesion wiring (not shown; see text); B1 at full capacity strongly outperforms both baselines ($g > 3$).
(b) Dual-task cost summary: per-step accuracy, report-window decision accuracy, and conjunction success by capacity. Conjunction collapses at bus-off; the report-window decision is most sensitive to capacity reduction.
(c) Workspace cue entry and displacement under backward masking. Cue signal (y-axis) jumps at cue entry, persists for a capacity-dependent duration, then decays as distractors displace cue content. Four slots: signal persists through 4 masks; two slots: faster displacement; bus-off: no signal (flat at 0).
Note: A finer 5-level capacity sweep (0\%, 25\%, 50\%, 75\%, 100\%) was conducted for the $\Delta$NRS analysis (see text) but is not shown here.
}
\label{fig:e2_results}
\end{figure}

Figure~\ref{fig:e2_results} presents the main findings across neural and behavioral measures.

\paragraph{Behavioral Performance Scales with Capacity.}
Conjunction accuracy showed a monotonic relationship with workspace capacity (Figure~\ref{fig:e2_results}a). Across $n = 20$ random seeds (32 evaluation episodes per seed), at full capacity (4 slots), B1 achieved $99.5\% \pm 2.0\%$ conjunction accuracy (mean $\pm$ SD across seeds). At reduced capacity (2 slots), performance dropped to $28.1\% \pm 9.7\%$, showing substantial seed-to-seed variability but a clear degradation. At bus-off (0 slots), conjunction accuracy collapsed to $0.0\% \pm 0.0\%$; the task became unsolvable without workspace access to cue information.

\textbf{Statistical validation:} The unit of analysis for all statistical tests is the random seed ($n=20$ per condition), treating each independently trained agent as a single observation (each seed contributes its mean over 32 evaluation episodes). We acknowledge that even this sample size limits power for moderate effects and yields nontrivial uncertainty around estimated thresholds. Concretely, with $n=20$ per group, a two-sample comparison is still underpowered for modest effects: a standardized effect on the order of Cohen's $d \approx 0.9$ is typically required to reach 80\% power at $\alpha=0.05$ under roughly Gaussian seed-to-seed variability. Accordingly, we interpret the strongest evidence as qualitative, seed-consistent regime changes (e.g., near-ceiling vs.\,collapsed-to-zero performance) rather than precise estimates of thresholds or effect sizes; we report uncertainty for transparency and treat effect sizes as descriptive. For transparency, we also report $p$-values in coarse scientific notation (one significant digit) and interpret them as descriptive rather than as precise evidence of a critical threshold. The difference between full capacity and half capacity was statistically significant (Welch's $t$-test: $t = 31.3$, $p < 10^{-16}$; Hedges' $g = 9.70$, large effect). A non-parametric permutation test confirmed the result ($p < 10^{-4}$). Comparison against baseline agents was also decisive: B1 at full capacity (per-step accuracy $90.7\% \pm 3.5\%$) outperformed A0 (16.1\%; $p < 10^{-9}$, $g = 3.30$) and A1 (15.2\%; $p < 10^{-9}$, $g = 3.54$). The similar near-chance performance of A0 and A1 is expected under strong-lesion wiring: because cues are routed exclusively through the workspace bus (which neither baseline has), both agents are effectively guessing regardless of whether they have recurrent memory.

The 4-slot workspace is analogous to classic findings on human visual working memory capacity \citep{luck1997capacity, cowan2001magical}, which suggest a limit of approximately 3--4 items. Our results demonstrate that reducing this capacity has behavioral consequences: with only 2 slots available, agents frequently failed to maintain both cues across the delay period, mirroring the interference effects observed when humans are asked to hold more items than their capacity allows.

\paragraph{Bus-Off Discontinuity in an Access Marker.}
Beyond the linear scaling of performance, we analyzed the no-report signature ($\Delta$NRS; Methods) across a 5-level capacity sweep (\texttt{workspace\_capacity\_scale} $\in \{0, 0.25, 0.5, 0.75, 1.0\}$; $n=5$ seeds). Across the tested settings, all tested non-zero capacities (25--100\%) produced strong above-baseline decodability ($\Delta$NRS $\approx 0.51$), while the complete workspace lesion (bus-off; 0\%) collapsed to near-zero / below-baseline decoding ($\Delta$NRS $\approx -0.12$). Given the coarse, discrete nature of this sweep, we interpret this as a qualitative discontinuity at a complete lesion (bus-on vs. bus-off), not as evidence for a precisely estimated threshold or for continuous thresholding dynamics. We do not claim critical scaling, a thermodynamic-limit phase transition, or a precisely estimated critical point; rather, the pattern indicates that in this architecture and task, access-related dynamics are robust to partial capacity reductions but fail catastrophically when the workspace is removed.

\paragraph{Dual-Task Cost Reveals Decision-Specific Demands.}
Figure~\ref{fig:e2_results}b summarizes performance at three granularities: per-step accuracy, report-window decision accuracy, and conjunction success. At full capacity, B1 achieves high per-step accuracy ($90.7\% \pm 3.5\%$) and report-window decision accuracy ($89.2\% \pm 3.6\%$), yielding near-ceiling conjunction accuracy ($99.5\% \pm 2.0\%$). At half capacity, per-step accuracy degrades ($44.8\% \pm 16.6\%$) and report-window decision accuracy is low ($19.6\% \pm 6.8\%$), producing only $28.1\% \pm 9.7\%$ conjunction accuracy. At bus-off, per-step accuracy is low ($12.7\% \pm 5.0\%$), conjunction accuracy collapses to $0.0\% \pm 0.0\%$, and report-window decision accuracy is low ($14.1\% \pm 6.4\%$), consistent with the absence of a functional access pathway for cue information. For interpretation, report-window decision accuracy has a $1/7 \approx 14.3\%$ chance floor (7 discrete actions), whereas conjunction accuracy requires both reports to be correct (chance $\approx (1/7)^2 \approx 2.0\%$ under independent guessing); thus, ``0.0\%'' indicates that no successful conjunction episodes were observed rather than implying a 0\% per-decision chance rate.

\paragraph{Cue Signal Dynamics Under Backward Masking.}
To visualize workspace dynamics at finer temporal resolution, we employed a backward masking paradigm (Figure~\ref{fig:e2_results}c), a technique widely used in consciousness research to probe the temporal dynamics of conscious access \citep{dehaene2014consciousness}. In visual backward masking, a brief target stimulus is followed by a mask that interferes with its conscious perception; here, we present distracting stimuli after the cue to test how long the cue representation persists in the workspace before being displaced.

We utilized a ``Report-Flag Signal'' metric, exploiting the finding that the \texttt{report\_flag} component creates strong separation in slot space (cosine similarity $\sim 0.22$ vs. $\sim 0.98$). This allowed us to construct a cue-specific projection that is positive when the cue is present and negative when displaced.

The results revealed capacity-dependent persistence of the cue signal:
\begin{itemize}
    \item \textbf{4 slots (100\%):} Signal jumped from 0 to $\sim$0.78 at cue entry, remained positive through 4 mask steps, then decayed to negative values ($\sim -0.26$) as slot 0 was overwritten.
    \item \textbf{2 slots (50\%):} Identical initial signal ($\sim$0.78), but faster decay; signal became negative by mask step 2 ($\sim -0.41$).
    \item \textbf{Bus-off (0\%):} Signal remained at 0 throughout (no workspace storage).
\end{itemize}

This pattern is analogous to human backward masking experiments: in both cases, the initial encoding of the target is identical across conditions (the $\sim$0.78 signal at cue entry), but the \emph{duration} of conscious access differs based on capacity constraints. Just as a stronger mask or shorter stimulus-onset asynchrony (SOA) reduces conscious report in humans, reduced workspace capacity shortens the window during which information remains accessible for report in our agents. This demonstrates the GWT prediction: workspace capacity determines how long task-relevant information persists before displacement by incoming stimuli.

\paragraph{Neural Signatures Track Capacity.}
GBI at full capacity averaged 0.66, indicating broad cross-slot coupling (high participation across slots) consistent with the ``global broadcast'' intuition in GWT. At half capacity, GBI dropped to 0.03 (reduced broadcast breadth; lower cross-slot participation), reflecting the diminished opportunity for information to be shared across the bus with fewer slots available. Ignition sharpness showed the expected pattern: IS $\approx 0.15$ at full capacity (bootstrap 95\% CI [0.06, 0.26] across $n=20$ seeds), IS $\approx 0.06$ at half capacity (bootstrap 95\% CI [0.02, 0.10]), and IS $= 0$ at bus-off (no ignition without a workspace).\footnote{We note that bootstrap confidence intervals can become narrow when seed-to-seed variance is low. These narrow CIs reflect genuine consistency across seeds rather than high precision about population-level effects; we report them for transparency but caution against over-interpreting their width.} The complete absence of ignition at bus-off demonstrates that ignition is not merely correlated with workspace activity but \emph{depends} on it. Without the workspace, there is no substrate for the non-linear amplification that characterizes conscious access in GWT.

\paragraph{Bus Audit Confirms Causal Dependence.}
The ``bus audit'' (see Methods) confirmed strong gradients through the bus pathway (T1 median Jacobian norms $\sim 78$--175) and zero leakage through the trunk (T3 trunk isolation 0\% flips), validating that behavior genuinely depended on workspace contents.

Given that metacognitive access depends on a Self-Model and information access depends on capacity, we now ask whether these indicators predict functional robustness to internal noise.

\subsection{Experiment 3 (Triangulation): Hierarchical Robustness and Composite Markers}

\subsubsection{Motivation}
Theories of consciousness are often treated as competitors, each proposing a unique neural signature. However, our previous experiments suggest they describe distinct functional layers. We selected this final angle to test the \textbf{Integration} and \textbf{Functional Utility} dimensions: linking abstract neural markers (ignition, complexity, calibration) to the concrete evolutionary advantage of robustness. By bridging this ``explanatory gap,'' we aim to show what consciousness is \emph{for}.

In Experiment~3, we test a \textit{Triangulation Hypothesis}: that a reduced composite of cross-architecture markers (GBI + $\Delta$PCI) predicts robustness better than either term alone, and that adding a HOT-aligned metacognition term (Type-2 AUROC) provides incremental information \emph{within} Self-Model agents. We explicitly evaluate whether these markers capture redundant information or contribute unique variance to robustness.

\subsubsection{Results}

\begin{figure}[htbp]
\centering
\begin{tabular}{cc}
\includegraphics[width=0.45\textwidth]{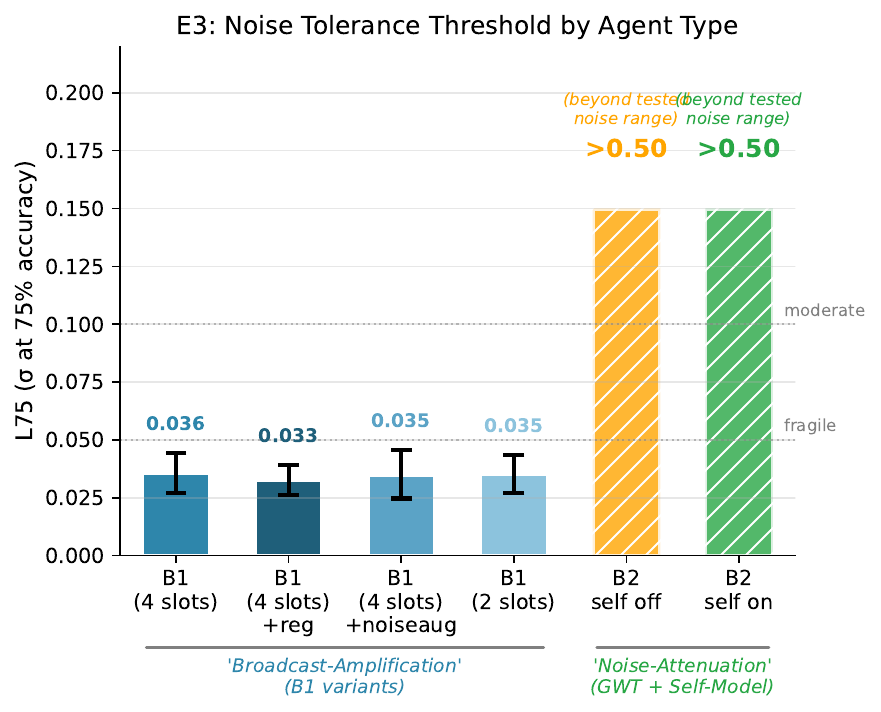} &
\includegraphics[width=0.45\textwidth]{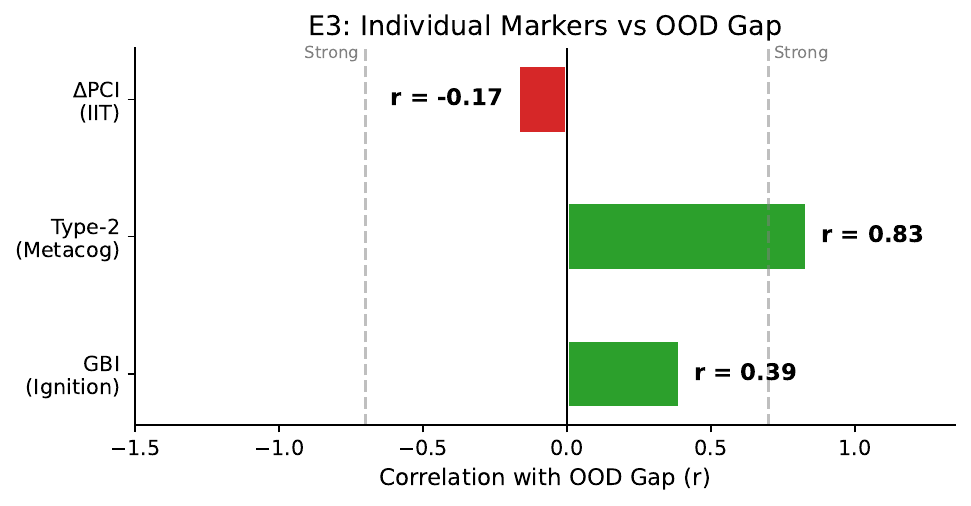} \\
(a) & (b) \\
\multicolumn{2}{c}{\includegraphics[width=0.7\textwidth]{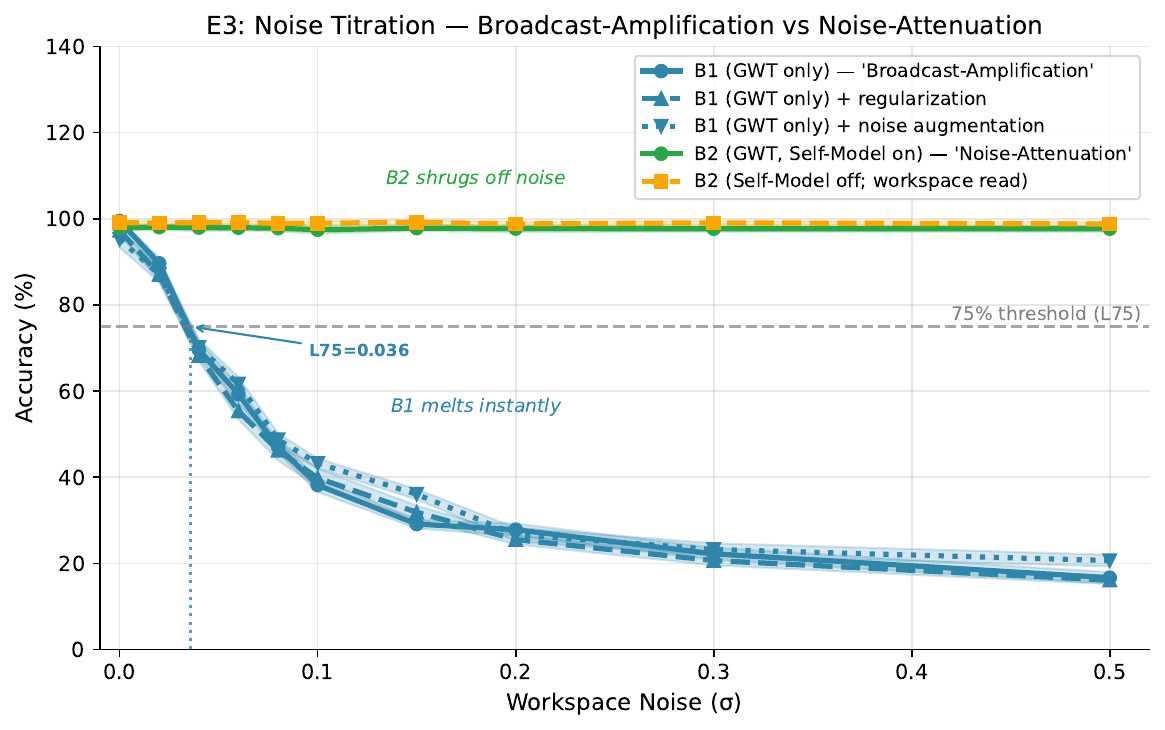}} \\
\multicolumn{2}{c}{(c)}
\end{tabular}
\caption{\textbf{Two co-equal findings: a PCI transfer constraint and a robustness hierarchy.}
(a) \textbf{Broadcast-Amplification vs.\ Noise-Attenuation:} B1 (GWT-only) collapses under minimal latent noise (L75 $\approx 0.04$); two steelman B1 variants with standard regularization or training-time noise augmentation remain similarly fragile (L75 $\approx 0.03$--$0.04$). B2 remains robust to latent noise within the tested range (L75 $> 0.50$, i.e., accuracy never dropped to 75\% for $\sigma \le 0.50$) both with the Self-Model enabled and in a Self-Model-off / workspace-read control. This suggests that, in this implementation, the robustness gain is not solely explained by routing actions through $z_{\text{self}}$.
(b) \textbf{Marker Correlations with OOD Gap:} Pearson correlations between individual markers and the OOD generalization gap (ID decision accuracy minus OOD decision accuracy; larger gaps indicate worse robustness). For agents without a workspace (and for the bus-off lesion), we define GBI as 0 ("no broadcast") for this correlation. The IIT-adjacent term is $\Delta$PCI (correct minus incorrect PCI-A; Methods), not raw PCI-A; raw PCI-A is reported separately and is inverted under the workspace bottleneck (Table~\ref{tab:pcia_inversion}).
(c) \textbf{Noise Titration Curves:} Performance degradation profiles for baselines, GWT (fragile), and Self-Model (robust) agents under latent perturbation.
}
\label{fig:e3_results}
\end{figure}

Experiment~3 yields two co-equal findings: (1) a negative result constraining the transfer of PCI-style complexity proxies to engineered agents, and (2) a robustness hierarchy distinguishing broadcast-only architectures from those with higher-order monitoring. We present the PCI transfer constraint first because it carries direct implications for IIT-adjacent assessment in artificial systems.

\paragraph{Finding 1: Raw PCI-A Inverts Under a Workspace Bottleneck: A Transfer Constraint for IIT-Adjacent Proxies.}
Table~\ref{tab:pcia_inversion} reports the \emph{raw} perturbational complexity (PCI-A; see Methods) across baseline and workspace agents. Contrary to a naive IIT/PCI transfer expectation (``more integrated'' $\Rightarrow$ higher perturbation-evoked trajectory complexity), the workspace agent showed \emph{lower} algorithmic complexity than both baselines (A0/A1). This is not a small or ambiguous effect; it is an explicit inversion that functions as a genuine empirical constraint on how PCI-style proxies behave in engineered systems.

\begin{table}[t]
\centering
\caption{\textbf{Raw PCI-A is inverted under a workspace bottleneck.} Mean $\pm$ SD across $n=20$ seeds. Higher values indicate higher Lempel--Ziv compressibility-based trajectory complexity.}
\label{tab:pcia_inversion}
\begin{tabular}{lc}
\toprule
\textbf{Architecture} & \textbf{Raw PCI-A} \\
\midrule
A0 (Feedforward baseline) & $0.84 \pm 0.18$ \\
A1 (Recurrent baseline) & $0.78 \pm 0.02$ \\
B1 (Workspace / GWT agent) & $0.33 \pm 0.02$ \\
\bottomrule
\end{tabular}
\end{table}

\textbf{Why does the workspace reduce trajectory complexity?} Mechanistically, our global workspace is a shared bottleneck that many downstream computations depend on. That design can reduce apparent complexity in at least three (non-exclusive) ways. (i) \emph{Low-rank forcing / redundancy:} if many units are driven by a common broadcast state, the evoked trajectories become more redundant across dimensions and therefore more compressible. (ii) \emph{Manifold stabilization:} bottlenecks and normalization can push dynamics toward a small set of task-relevant attractor-like modes, so perturbations decay back toward a structured manifold rather than producing richly branching transients. (iii) \emph{Failure-driven ``complexity'' in baselines:} recurrent baselines can produce high Lempel--Ziv complexity through unstructured variability, especially when the perturbation pushes them off-policy; high compressibility-based complexity is not guaranteed to reflect meaningful integration.

\textbf{Is this a fundamental problem with PCI as a consciousness measure?} We interpret the result as \emph{implementation- and setting-specific}, not as a refutation of PCI or IIT. PCI in humans is designed around spatiotemporal complexity across many channels following a perturbation (e.g., TMS--EEG) in a system whose baseline dynamics include rich spontaneous activity. Our PCI-A proxy instead measures the compressibility of a particular internal trajectory under an engineered bottleneck and a synthetic perturbation scheme. The negative result therefore primarily cautions that ``absolute algorithmic complexity of a latent trajectory'' is not an architecture-invariant stand-in for integration when moved from biological cortex to task-trained artificial agents.

\textbf{Why use $\Delta$PCI at all?} We report $\Delta$PCI (correct minus incorrect trial complexity) as a contrastive \emph{selective complexity} diagnostic that partially controls for noise/failure confounds. In these agents, incorrect trials often reflect destabilized, off-policy behavior that can inflate compressibility-based complexity without corresponding to successful information processing. Conditioning on correctness treats behavioral success as an external anchor for ``meaningful'' processing and asks a narrower question: does perturbational complexity increase \emph{selectively} when the agent is in the regime that supports correct computation? We emphasize that $\Delta$PCI is not a canonical IIT prediction and should not be read as a principled surrogate for $\Phi$; it is a pragmatic control that helps separate \emph{structured} evoked dynamics from \emph{unstructured} failure-driven variability. In our hierarchical regression, its incremental contribution to OOD robustness is modest ($\Delta R^2 \approx 2\%$), and we interpret this as further evidence that PCI-style proxies require care when ported to engineered agents.

\paragraph{Finding 2: Broadcast-Amplification vs.\ Noise-Attenuation.}

\begin{figure}[htbp]
\centering
\includegraphics[width=0.95\textwidth]{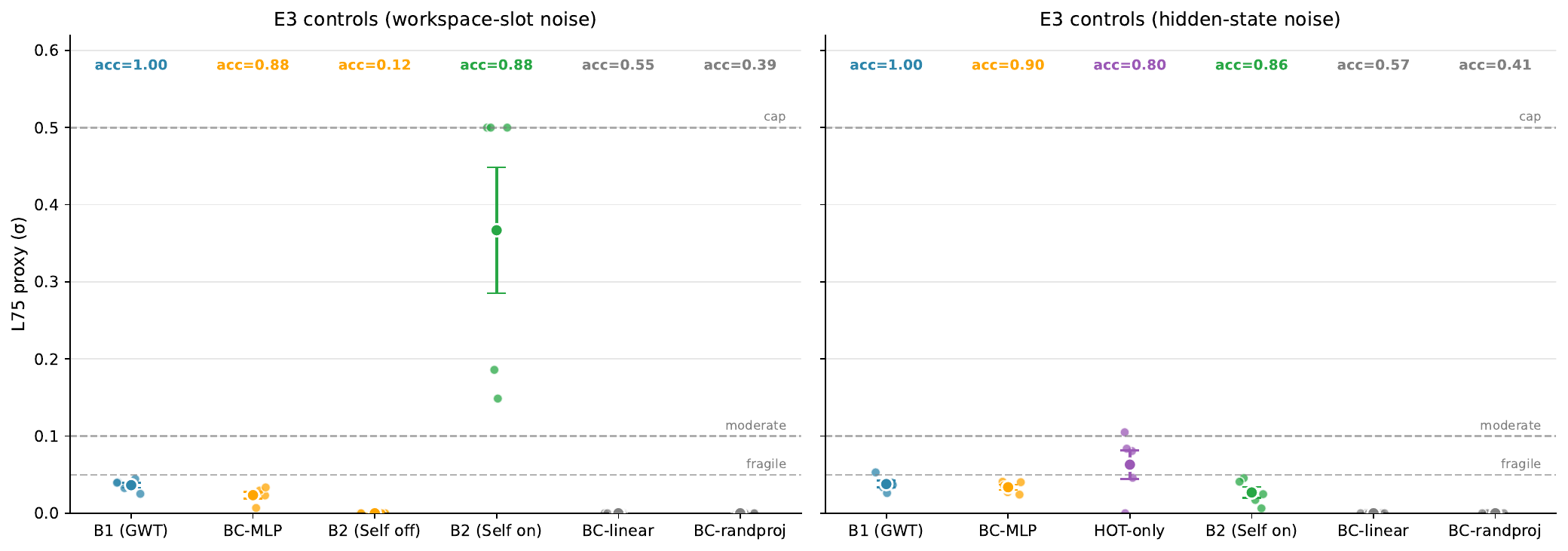}
\caption{\textbf{Denoising attribution controls (pilot; $n=5$ seeds).} Left: latent workspace-slot noise (the primary Experiment~3 perturbation) comparing B1 (GWT-only), B2 (Self-Model), and matched ``BC-only'' compressor controls (a 64-d bottleneck applied directly to the flattened workspace state). Right: latent hidden-state (post-GRU carrier) noise comparing the same workspace agents to a ``HOT-only'' agent (A1 + $z_{\text{self}}$ without workspace; cues visible because there is no workspace bus). Points are seeds; squares show mean $\pm$ SD for finite thresholds; triangles at $\sigma=0.5$ denote seeds that did not cross the 75\% threshold within the tested noise range.}
\label{fig:e3_denoise_controls}
\end{figure}
The noise titration revealed a dissociation between broadcast capacity and robustness (Figure~\ref{fig:e3_results}a). The GWT agent (B1), despite having high report-step decision accuracy (100\%) and high GBI (0.65), proved fragile, exhibiting broadcast-amplification with an L75 of $\sigma \approx 0.04$. To put this in perspective: the workspace activations have unit variance after normalization, so $\sigma = 0.04$ represents noise at just 4\% of the signal magnitude. Yet this tiny perturbation was sufficient to drop performance from 100\% to 75\%. By $\sigma = 0.10$ (10\% noise), accuracy had collapsed to near chance.

To address the concern that this is a straw-man baseline, we repeated the titration on B1 with two standard robustness interventions: (i) dropout + weight decay during behavior cloning, and (ii) training-time latent workspace-noise augmentation. Neither materially improved noise tolerance (L75 $= 0.033 \pm 0.007$ and $0.035 \pm 0.010$, respectively; $n = 20$ seeds).

This extreme sensitivity arises because B1's policy reads directly from the workspace slots. Any noise injected into the workspace propagates directly to the action selection, with no opportunity for filtering or error correction. The workspace acts as an amplifier: it broadcasts information globally, but it also broadcasts noise globally.

In contrast, the Self-Model agent (B2) demonstrated robustness beyond the tested noise range (L75 $> 0.50$; accuracy never dropped to 75\% for $\sigma \le 0.50$), exhibiting noise-attenuation. Even at $\sigma = 0.50$ (noise equal to 50\% of signal magnitude), B2 maintained $>$80\% accuracy. This robustness persisted in the Self-Model-off / workspace-read control (Figure~\ref{fig:e3_results}a), indicating that in this implementation robustness is not strictly dependent on routing actions through $z_{\text{self}}$ at inference. While $z_{\text{self}}$ is a strongly compressed summary of the workspace state (effective dimension $\sim$2), the noise-tolerance advantage appears to be a broader property of the B2 training/architecture rather than a bottleneck-only effect.

\paragraph{Denoising attribution baselines.}
A key alternative hypothesis is that the robustness gain in Figure~\ref{fig:e3_results} is not specific to the Self-Model, but would emerge from adding any low-dimensional bottleneck on the action pathway. To test this, we evaluated (i) a ``HOT-only'' agent (A1 + $z_{\text{self}}$ without a workspace bus) and (ii) matched ``BC-only'' compressor controls that replace $z_{\text{self}}$ with a 64-d bottleneck on the flattened workspace state. In a pilot sweep ($n=5$ seeds; Figure~\ref{fig:e3_denoise_controls}), the learned MLP bottleneck achieves high baseline accuracy ($0.879 \pm 0.054$) but remains fragile under workspace-slot noise (L75 $= 0.023 \pm 0.010$), far below the Self-Model agent (3/5 seeds remain above the 75\% threshold at $\sigma=0.5$, and the remaining seeds yield L75 $= 0.167 \pm 0.026$). Linear and random-projection bottlenecks underperform on baseline accuracy and do not yield meaningful robustness comparisons in this task. Under hidden-state noise, the HOT-only agent (evaluated with cues visible, since it has no workspace bus) maintains baseline accuracy $0.799 \pm 0.169$ and shows improved tolerance (L75 $= 0.063 \pm 0.041$), indicating that a Self-Model bottleneck can also stabilize recurrence-only computation, but that the strongest robustness effect in this study remains specific to the workspace-noise regime.

\paragraph{Composite Predicts OOD Fragility Across Architectures.}
In the canonical OOD-gap analysis (Figure~\ref{fig:e3_results}b), we correlate individual markers with the OOD generalization gap (ID accuracy minus OOD accuracy; larger gaps indicate worse generalization). The \emph{positive} correlations indicate that higher marker values are associated with \emph{larger} OOD gaps (i.e., worse OOD generalization). This is distinct from L75 noise tolerance, where B2 excels; OOD fragility reflects a different failure mode (distribution shift rather than latent noise). Broadcasting alone has modest predictive power: GBI yields Pearson $r \approx 0.39$ (equivalently $R^2 \approx 0.15$). Adding $\Delta$PCI provides only a small incremental gain ($\Delta R^2 \approx 2\%$). Adding Type-2 AUROC improves in-sample fit (Pearson $r \approx 0.79$; leave-one-out $r_{\text{LOO}} \approx 0.62$), but AUROC is near-constant for non--Self-Model agents, so this term partly encodes the Self-Model design choice.

\paragraph{Non-Circularity Controls.}
Because Type-2 AUROC is $\approx 0.50$ for agents lacking a Self-Model, we treat AUROC-free composites (GBI + $\Delta$PCI) as the primary cross-architecture triangulation score and interpret AUROC gains as mechanism-specific rather than independent evidence. Accordingly, we treat the CTS$\to$L75 link as a \emph{mechanistic consistency check} within this synthetic setting rather than a discovery-grade predictive model; broader generalization would require additional architectures where AUROC varies continuously.

\paragraph{Unique Variance Decomposition.}
Hierarchical regression (Figure~\ref{fig:e3_results}c) revealed the distinct contributions of each layer. In hierarchical regression, predictors are added sequentially to determine how much additional variance each explains beyond what is already captured by earlier predictors. The $R^2$ value indicates the proportion of variance explained (ranging from 0 to 1, where 1 means perfect prediction), and $\Delta R^2$ indicates the unique contribution of each added predictor.
\begin{itemize}
    \item \textbf{Layer 1 (Broadcast):} GBI alone explains $\sim$15\% of the variance in OOD gap (Pearson $r \approx 0.39$). This confirms that broadcasting capacity matters, but leaves $\sim$85\% unexplained.
    \item \textbf{Layer 2 (Integration):} Adding $\Delta$PCI yields a small but non-zero gain ($\Delta R^2 \approx 2\%$), bringing the total to $\sim$17\%. This does \emph{not} overturn the inverted absolute PCI-A ordering (Table~\ref{tab:pcia_inversion}); instead it reframes the question contrastively: does perturbational complexity increase \emph{selectively} on successful trials (``meaningful processing'') relative to failed trials (``noise complexity'')?
    \item \textbf{Layer 3 (Metacognition):} Adding Type-2 AUROC yields a further gain ($\Delta R^2 \approx 19\%$), bringing the total to $\sim$37\%. However, as noted above, AUROC is near-constant (0.50) for all agents lacking a Self-Model, so this term primarily distinguishes B2 self-on from the rest. The key result is that \textbf{GBI + $\Delta$PCI alone explain $\sim$17\% of variance in OOD gap}, markers that are well-defined across all architectures.
\end{itemize}
This pattern is consistent with a hierarchical interpretation: broadcast capacity (GBI) and processing richness ($\Delta$PCI) provide the foundation, while Type-2 AUROC mainly distinguishes Self-Model agents; the robustness control in Figure~\ref{fig:e3_results}a suggests the robustness gain is not solely mediated by $z_{\text{self}}$ at inference. We caution against interpreting the AUROC contribution as independent evidence, given its structural confound with the Self-Model design.

\section{General Discussion}

\subsection{Summary of Findings}
We introduced a synthetic neuro-phenomenology framework to stress-test consciousness indicators. Our experiments reveal a hierarchical functional architecture defined by both capabilities and surprising failure modes:

\textbf{Experiment~1} showed a double dissociation in this agent family: Self-Model ablation abolishes metacognitive calibration ("knowing") while leaving first-order performance ("doing") largely intact.

\textbf{Experiment~2} provides evidence that workspace capacity is causally necessary for information access within this setup ($n = 20$ seeds; full vs. half capacity: $p < 10^{-4}$, $g = 1.65$; B1 vs. A0/A1: $p < 10^{-9}$, $g > 3$), and reveals a sharp discontinuity at the complete workspace lesion (bus-off) across the tested discrete capacity settings. When the workspace is fully lesioned (bus-off), the access-related dynamics collapse entirely; with $n=20$ seeds and discrete capacity levels, we treat this as a qualitative regime contrast rather than a precisely estimated threshold.

\textbf{Experiment~3} revealed the \textbf{broadcast-amplification} effect: the global workspace broadcasts information but amplifies noise, creating extreme fragility. B2 agents remain robust in the same perturbation regime (including in a Self-Model-off / workspace-read control), yielding a \textbf{noise-attenuation} pattern relative to B1. Experiment~3 also produced an explicit negative result for a naive IIT/PCI-style complexity proxy: raw PCI-A is inverted under the workspace bottleneck, and $\Delta$PCI contributes only modest unique variance.

\subsection{Indicators vs. Pseudo-Consciousness}
Our agents are not conscious; they are reference implementations of theories. The fact that B2 passes HOT-style Self-Model tests and B1 passes GWT tests without ``feeling'' anything highlights the risk of pseudo-consciousness. However, the \emph{functional} properties revealed (especially the robustness of the B2 family under latent noise perturbations) suggest that these architectural features provide tangible evolutionary advantages beyond mere ``inner light.''

\subsection{Hierarchies of Robustness and their Limits}
The dissociation between B1 (broadcast-amplification) and B2 (noise-attenuation) challenges the assumption that broadcasting equals robustness. The global workspace amplifies signals, but without a higher-order monitor, it also amplifies noise. This suggests that GWT and Self-Model mechanisms co-evolved: workspace for rapid dissemination, Self-Model for quality control.

However, the Self-Model is not a "homunculus" that cares about truth; it is a mechanism optimized for reward. This implies that introspective access is not an automatic byproduct of having a Self-Model; it emerges from the architectural constraint that forces behavior to be mediated by the compressed self-representation.

\subsection{A Constraint on PCI-Style Transfer to Engineered Agents}
The inversion of raw PCI-A under the workspace bottleneck (Table~\ref{tab:pcia_inversion}) is one of the most consequential findings for the IIT community. Naive transfer of PCI-style measures (where ``more integrated'' architectures are expected to show higher perturbation-evoked trajectory complexity) fails in our engineered agents. This is not a methodological nuisance but a genuine empirical constraint: in systems with shared bottlenecks and normalization, tighter coupling can \emph{reduce} apparent complexity by forcing dynamics onto lower-dimensional manifolds and increasing redundancy across channels. Meanwhile, unstructured failure dynamics in less-organized systems can inflate Lempel--Ziv complexity without reflecting meaningful integration.

This result does not refute PCI or IIT as applied to biological systems, where the measure was designed for spatiotemporal complexity across many channels following TMS perturbations in a brain with rich spontaneous activity. Rather, it cautions that absolute algorithmic complexity of a latent trajectory is not architecture-invariant when moved from biological cortex to task-trained artificial agents. Future work on AI consciousness assessment should explicitly validate which perturbation and readout choices recover IIT-aligned monotonicities before treating PCI-style proxies as universal integration measures.

\subsection{Implications for AI Assessment}
Our results support a triangulation approach to AI consciousness assessment: no single marker is sufficient, and different markers can fail in different ways. In our synthetic setting, a reduced composite of cross-architecture markers (GBI + $\Delta$PCI) predicts robustness, while metacognitive calibration (Type-2 AUROC) provides additional diagnostic power primarily \emph{within} Self-Model agents. This avoids treating AUROC as a universal continuous predictor in a regime where it is near-constant for non--Self-Model architectures. The core functional lesson remains: a system that ignites and broadcasts without quality control can be brittle (B1), whereas a system that filters and monitors its own broadcast stream is robust (B2).

This has practical implications for AI safety and alignment. An AI system that lacks metacognitive calibration cannot accurately report its own uncertainty, making it difficult for human operators to know when to trust its outputs. Our results suggest that metacognitive capacity is not merely a philosophical curiosity but a functional requirement for reliable AI systems. Future assessments should prioritize such multi-theory triangulation over the search for single canonical indicators, asking not just ``does this system broadcast information?'' but also ``does this system know what it knows?''

\subsection{Limitations and Future Work}
\label{sec:limitations_future}
Several limitations should be noted. First, we report statistics across $n=20$ training seeds per condition for the main comparisons in Experiments~1--3 (seed as the unit of analysis). Some targeted sweeps and attribution controls are pilots ($n=5$ seeds) and should be interpreted accordingly. This sample size is sufficient to demonstrate large qualitative regime changes, but it remains underpowered for moderate effects and yields uncertainty around estimated thresholds; wide intervals in intermediate regimes (e.g., reduced capacity) reflect genuine seed-to-seed variability rather than mere measurement noise. We did not run a prospective power analysis, and we treat quantitative effect sizes and estimated thresholds as exploratory proof-of-concept results within this specific architecture family and task suite.

Second, our IIT-adjacent complexity probe produced an explicit negative result for a naive PCI-style transfer prediction: raw PCI-A was inverted for workspace agents (Table~\ref{tab:pcia_inversion}). We argue this inversion is informative rather than embarrassing: in engineered bottlenecked systems, tighter coupling and organization can reduce compressibility-based trajectory complexity, while unstructured failure dynamics can inflate it. While $\Delta$PCI provides a modest contrastive ``selective complexity'' signal, it remains a heuristic diagnostic (not a direct test of $\Phi$) whose behavior depends on how failure/noise contributes to trajectories. Future work should implement more direct approximations to $\Phi$ and/or alternative whole-vs-parts integration measures that are better matched to software agents, and should explicitly test which perturbation and readout choices recover the intended IIT-aligned monotonicities.

Third, our tasks were relatively simple gridworlds with discrete actions and limited state spaces. The broadcast-amplification / noise-attenuation distinction may manifest differently in richer, continuous-control environments where noise can accumulate over longer time horizons.

\textbf{Denoising attribution controls (scope).} We include HOT-only and BC-only compressor controls (Figure~\ref{fig:e3_denoise_controls}) to tighten attribution for the Experiment~3 robustness result. However, these controls were run as a pilot ($n=5$ seeds) and span a limited compressor family (linear projection, small MLP, fixed random projection). Some compressor baselines also underperform on baseline accuracy, which limits the interpretability of their robustness thresholds. Future work should expand these attribution controls (more seeds; parameter-matched variants; additional bottleneck placements; and stronger learned compressors such as autoencoder-style reconstruction objectives) to test whether conclusions remain stable under broader architectural variation.

Fourth, our agents are trained via behavior cloning from an expert oracle, and Experiment~1 includes an explicit auxiliary loss supervising the confidence head. This raises a potential concern that the Self-Model could merely learn to predict the oracle's behavior rather than monitor the agent's own epistemic state. Two clarifications are important. (i) The oracle is used only to generate training targets; it is never consulted at inference, and all reported effects are measured in autonomous rollouts of trained agents under \emph{post-training} perturbations (lesions, capacity reduction, and latent workspace noise) that are out-of-distribution relative to the demonstrations. (ii) The metacognitive supervision target is defined by the agent's own intended report decision being correct (argmax of the agent logits compared to the oracle-provided ground-truth label at that step). Thus, the Self-Model is trained to answer ``will \emph{my} current decision be correct given my internal state?'' rather than to predict whether the oracle would be correct. Nevertheless, reinforcement learning can interact with the same architectures in qualitatively different ways (exploration, credit assignment, and stability), and establishing whether comparable dissociations emerge under RL (particularly whether calibrated metacognition can be learned from reward alone without labeled correctness) remains an open and important direction for future work.

Finally, our agents are not conscious in any phenomenological sense; they are reference implementations of theories. The fact that B2 passes HOT tests without ``feeling'' anything underscores the gap between functional indicators and subjective experience. However, we argue that functional validation is a necessary (if not sufficient) step toward understanding consciousness: if a proposed mechanism does not produce the predicted functional consequences even in a simplified synthetic system, it is unlikely to be the correct account of the biological phenomenon.

Future work will investigate how these architectural components interact under greater temporal conflict, test whether the hierarchical relationship between broadcast and metacognition generalizes to language models and other modern AI architectures, and explore mechanisms for maintaining metacognitive stability without explicit supervision. Ultimately, the goal is to establish whether the triad of broadcast, integration, and metacognition serves as a robust predictor of general intelligence and adaptability across diverse domains.

\bibliographystyle{plainnat}
\bibliography{references}

\end{document}